# Evolutionary Algorithm for Sinhala to English Translation

J.K. Joseph[1], W.M.T. Chathurika[1], A. Nugaliyadde[1,2], Y. Mallawarachchi[1]

`jenny.k@sliit.lk,thilini.w@sliit.lk,a.nugaliyadde@murdoch.edu.au,yasas.m@sliit.lk`

[1]*Department of IT, Sri Lanka Institute of Information Technology, New Kandy Road, Malabe, Sri Lanka*
[2]*College of Science, Health, Engineering and Education, Murdoch University, Western Australia*

*Abstract*— **Machine Translation (MT) is an area in natural language processing, which focus on translating from one language to another. Many approaches ranging from statistical methods to deep learning approaches are used in order to achieve MT. However, these methods either require a large number of data or a clear understanding about the language. Sinhala language has less digital text which could be used to train a deep neural network. Furthermore, Sinhala has complex rules therefore, it is harder to create statistical rules in order to apply statistical methods in MT. This research focuses on Sinhala to English translation using an Evolutionary Algorithm (EA). EA is used to identifying the correct meaning of Sinhala text and to translate it to English. The Sinhala text is passed to identify the meaning in order to get the correct meaning of the sentence. With the use of the EA the translation is carried out. The translated text is passed on to grammatically correct the sentence. This has shown to achieve accurate results.**

## I. INTRODUCTION

With large amounts of context available in the internet in different languages, language translation has achieve more interest from an industrial and research perspectives. In order to handle the large stacks of text data Machine Translation (MT) was introduced. MT targets to use computational models in order to achieve meaningful translations from one language to another.

In order to develop a MT computational model it is required to understand both the language, or to have language specialist. Therefore, it is not always practical to develop computational models to develop computational models for MT. Computational MT models have evolved from simple rule based models to deep learning models [1]. Statistical models have shown to perform well in many natural language understanding tasks [2] [3]. Applying statistics require a clear understanding about the language and the language structure [4]. However, this requires a large amount of knowledge about the language, which can be hard to achieve on languages which are not popular, or which have complex structures [5].

Compared to other popular languages Sinhala language has a smaller digital footprint. Therefore, using Sinhala for natural language processing tasks are harder [2]. Furthermore, Sinhala has a complex language structure therefore, it requires Sinhala expert knowledge [6]. Therefore, applying computational models to Sinhala language can be a hard task. Natural language processing tasks for Sinhala language were conducted using many techniques.

The rule based approach was used in simple Sinhala natural language tasks [7] . Similarly MT for Sinhala was also done using rule based approaches [8, 9]. However, since language is a complex tasks, statistical models were created and tested for Sinhala. Statistical models were used for Sinhala natural language tasks. The statistical models shows a potential in achieving better natural language models for Sinhala. Statistical models were used for MT in Sinhala which have gained traction [10] [11]. Statistical models have been vastly used to create complex Sinhala language models. These models were used on many industrial language tasks since the models are better. However, statistical models fail to capture all features of language models [12]. Statistical models require hand selected features therefore, expert knowledge is required [13] [14].

Deep learning does not require any feature selections but is capable of achieving higher accurate results compared to statistical methods and rule based models [12, 15-17]. Deep learning has been applied for MT [18] [19]. A deep learning model however, requires larger dataset to learn. Therefore, Languages with smaller digital data presence fail in producing better results using deep learning [7]. Deep learning requirement to use large data enables the deep learning model to generalize [20] [21]. Applying deep learning for MT in Sinhala does not have a high potential of achieving comparative results.

In this paper an Evolutionary Algorithm (EA) is used for MT from Sinhala to English. EA is an algorithm which focuses on the evolutionary mechanism. EA has 4 steps: Initialization, selection, genetic operation and termination. These steps are similar to natural selection in the environment. Until final batch is created, selection and genetic operation iterates [22]. EA has the possibility of achieving natural language tasks [23]. This is one of the first studies that have used EA to achieve MT. Our model uses EA in order to translate the text and generate the final text using the evolution. Direct translation would generate the English text. EA would use the evolutionary iteration to align and correct the grammatical structure. This would generate the final proper meaningful translation. EA does not require a large training dataset enabling languages similar to Sinhala to be used in MT.

## II. RESEARCH METHODOLOGY[1]

Language model is a network it is contains with words or alphabets and it is used for the training purpose. Also this model holds the language vocabulary. That was created by using Cygwin software in order to generate binary format of Language model.

### B. MEANING IDENTIFICATION

Language translation is twisted together with the "Meaning Identification". The base of the translation is identifying the appropriate meaning. This goes along with the word sense disambiguation which arises due to the different meaning for one word [1]. According to the rest of the content of the sentence, the meaning of the word will differ. The hardest part is to identify the correct meaning of the word by considering the remaining parts of the sentence. Most of the researches failed due to the incorrect solutions of this meaning identification function. Therefore this function is well known as a crucial one which will always needed the most attention. In this research the Sinhala text will be the input. Therefore we have to identify the meaning of a Sinhala text which is even harder than the English texts. So far there are number of researches that have been conducted for English language. As a result they have found so many different ways to get the meaning of the text such as taggers, parsers etc [8].

There were few researches that have been conducted for the meaning identification in language translation for Sinhala. But the main problem that can be identify is the researches aim about the word by word translation. Although the researches aimed about word by word translation, to get an accurate result we have to consider the whole sentence and identify the meaning. When doing this meaning identification by considering the whole text instead of only one word, we get the relevant word and then consider about the remaining words of the sentence. Then we can get the meaning of the text as a whole which will lead to an accurate result [9]. As an example, if we have four words in a sentence and we want to get the meaning by considering one word at a time there will be a word by word translation. In that case we don't bother regarding the altogether meaning of the words and we consider about single words. But this will lead to lot of inaccurate results as a word alone can give a different meaning and the same word along with few words can give a different meaning. In order to perform the meaning identification task successfully the concept called Point-wise Mutual Information will be used.

By using PMI values, can get the most occurred meanings for a given word and can avoid the meanings which will never be there. The concept of the PMI is, when there are two words called x, y and if they have P(x) and P(y) the mutual information will be calculated by using the following formula.

$$PMI(X,Y) = \log_2 \frac{P(x,y)}{P(x)P(y)}$$

The main idea of the PMI value is to consider the probability of the x and y together along with the probabilities of the x and y separately. The answer of the above formula will be considered and if it's a high value it says that there is truly an association between x and y [13]. Furthermore the values from the PMI will be taken as Positive Point –wise Mutual Information (PPMI) by replacing the PMI values that are negative values with zero [11]. By that we can avoid the unnecessary mistakes of language translation. This research was conducted in the Windows environment by using Python as the language. Python is a mostly used high level language which supports multiple programming paradigms. NLTK is a free and open source platform for Python programs which helps to work with human languages.

A word corpus is being built in the memory by calculating the relevant PPMI values. Then in the meaning identification process this word corpus is used as an aid. In this word corpus there are texts along with the PPMI values for each texts. Therefore when a text is given as an input for the meaning identification function, the word corpus that we have built is used to identify the meanings by considering the whole text and at the same time we can avoid unnecessary mistakes as we can identify the meanings that can never occur with the rest of the content of a text. This will lead to get an accurate result in this meaning identification function.

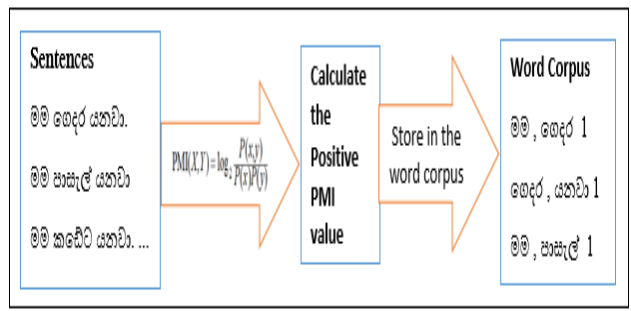

### C. TRANSLATION MODULE

For the translation phase, Meaning identified Sinhala sentence or a paragraph is an input. System split sentences and do the tokenization, System leads to do dictionary lookup and lexical analyzing using Sinhala to English dictionary. System need to catch unknown words, digits, signs etc. Then the system will morphologically analyze the tokens and identify the subject-predicate of the sentence. After that system will synthesis and morphologically process the target language (English). Finally rearrange the sentence to subject-verb-object format using POS tagging (Part Of Speech Tagging) [1]. Discussing about the direct or literal translation process, it is a dictionary based machine translation used which is based on entries of the language dictionary. The first generation of machine translation was entirely based on machine readable or electronic dictionaries. This methods is still helpful in translation [12]. The final output of the direct translation phase is not a complete intermediary sentence structure. A word-for-word (literal) translation occurs in this phase. Translation proceeds in a number of steps, each step dedicated to a specific task. The most important component is the bilingual dictionary. In this method, the Source Language text is structurally analyzed up to the morphological level, and matched for a particular pair of source and target language. This process of this system depends on the quality and quantity of the source-target language dictionaries, In order to achieve higher accuracy level,"Arutha" Dictionary used about seventy two thousand distinct words.

---
[1] Once accepted the code will be shared for public access with the text.

Commonly Machine Translation systems are bilingual (designed for two particular languages). But there are multilingual systems as well (designed for more than two languages). A bilingual systems can be either unidirectional or bidirectional. Usually majority of bilingual systems are unidirectional which translates from one Source Language (SL) to one Target Language (TL). Multilingual systems are always bidirectional. Ambiguity of a natural language is a major barrier for developing a quality translator. There are two types of ambiguity occurred while doing translation called Structural ambiguity and Lexical ambiguity [10].

The grammatical rules of Sinhala and English languages are different. In English language the sentence formation follows Subject Verb Object (SVO) where are in Sinhala the sentence formation is in the form of Subject Object Verb (SOV). Differences of Sinhala and English sentence formation patterns causes to grammar errors when it comes to direct translation.

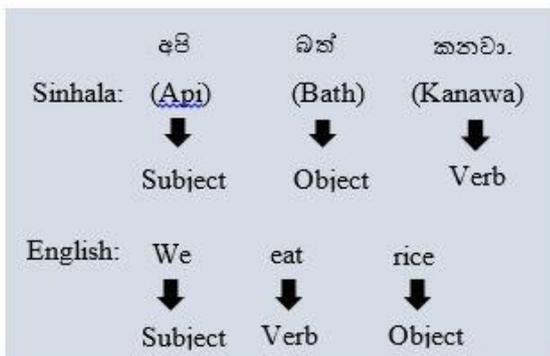

To create a dictionary of Sinhala and English fonts in the system we have to go through the following process.

Download the appropriate Sinhala fonts in your system then copy .ttf file and paste the file in the **FONTS** folder which is available in the **Control Panel** of the PC. There are two ways to type the Sinhala text. First method can used by the people who know the keyboard format of Sinhala fonts. After adding the fonts we need to go to **Control Panel** -**Region** – **Languages preferences** – add Sinhala - Ok. Following method is a simple method to type Sinhala.

We need to type the text in English which will directly convert the text into Sinhala. To achieve this process we need to download Google input tool. Need to create Sinhala and its equivalent English words text dictionary.

Meaning identified Sinhala sentence or a paragraph coming from the meaning identification function is an input for the direct translation function. After receiving the text, the sentence has been separated into words based on the delimiters (space, commas, full stops etc.).The separated words are stored in an Array List. From the Array List each and every word will check with the dictionary file and translate each and every word to English directly.

*D. EVOLUTIONARY ALGORITHM & GRAMMAR*

The research aim is to translate Sinhala to English through the meaning identification by using evolutionary algorithm. Evolutionary algorithm used to get grammatically correct sentences. EA uses selection and variation as basic principles. Competition among living beings is represented by Selection, to survive and reproduce their genetic information some should be better than others, natural selection is represented by this process. Each solution is given a change to reproduce and the quality is assessed by the fitness factor of each context. Variation shows the capacity of creating new living beings by using mutation. Likewise words will be selected, varied and create best quality sentences using Evolutionary Algorithm.

BEGIN
    INITIALISE population with unordered words;
    EVALUATE the candidate;
    REPEAT UNTIL (TERMINATION CONDITION is satisfied) DO
        SELECT parent;
        EVALUATE the parent;
        MUTATE by randomly pasting words;
        EVALUATE each child;
        SELECT fittest individual for next generation;
    OD
END

The Common idea about the evolutionary algorithm is same. When a population is given, in the environment only the fittest will survive and it cause to increase the fittest one of the population and increase quality of the population. The fitness function is a requirement to fulfill the adaptation to the system. It helps to get an accurate result therefore it is a measure of improvement. Based on the fitness of the sentence some better sentences are chosen to seed the next generation by using mutation factor. Mutate factor randomly paste words in order to gain new sentences. This process can be iterated until find a sufficient fittest and grammatical fluent sentence is created. Variation operation creates the necessary diversity and selection act as a force that improves quality. The combination of variance and selection improve the fitness value of the sentence. It is easy to good to see a optimize value or at least approximate value by approaching optimal values closer and closer over its process. Every time the process runs, it makes the population adapt to the environment better and better.

```
NP:    {<PRP>?<JJ.*>*<NN.*>+}
CP:    {<JJR|JJS>}
VERB:  {<VB.*>}
THAN:  {<IN>}
COMP:  {<DT>?<NP><RB>?<VERB><DT>?<CP><THAN><DT>?<NP>}
```

As above mentioned this process will create much accurate result, it begins after identifying the grammar of that particular sentence. Firstly grammatically incorrect sentence is taken into a list and by using nltk libraries sentence is broke into words and tagged as verbs, nouns etc. By using nltk library a key and a value is given to each sentence in order to identify the words separately. By using the key we can access the words, it helps when placing the words in correct grammatically form. Then created context is transfer to the algorithm in order to check fitness. In fitness function calculates the distance between target and the candidate sentences and check whether it close to the target. If it is equal to zero the expected output has reached. Else a sentence which is closest to the expected target is taken and mutates it. Random number is created and that number is used as the distance between two words. And words are randomly moved according to that number. 100 words are created at once. And this process is going until while (fitness < 0 and count < 1000) condition fulfill. By setting count's maximum value to 1000, CPU utilization is limited; otherwise some processed might go up to infinite times. Then the fitness of created sentences are checked, highest fitness is taken and it is taken as the parent of next generation. This process is iterated until it's reached for zero.

III. CONCLUSION

This paper introduces a model which uses EA to apply for MT. The proposed model translates English to Sinhala. However, a similar model can be applied to languages which have minimal digital text. The model directly translates Sinhala text to English. The direct translation would not hold the correct meaning of the Sinhala text. Using EA translated text is converted into a meaningful English text. Evolutionary process would continue until the text is grammatically correct. Furthermore, this study shows the possibility of applying EA to languages which have less digital text. Therefore, without moving towards deep learning, it is possible to achieve MT without training data, with high accuracy.